\documentclass[letterpaper, 10 pt, conference]{ieeeconf}  
\pdfoutput=1    

\IEEEoverridecommandlockouts                              

\usepackage{cite}
\usepackage{amsmath,amssymb,amsfonts}
\usepackage{algorithmic}
\usepackage{graphicx}
\usepackage{textcomp}
\usepackage{xcolor}
\usepackage{multirow}
\usepackage[hidelinks]{hyperref}

\usepackage{eso-pic}
\newcommand{\mycopyrighttext}{%
  \footnotesize
  \noindent
  \textcopyright~2025 IEEE. Personal use of this material is permitted.
  Permission from IEEE must be obtained for all other uses, in any current
  or future media, including reprinting/republishing this material for
  advertising or promotional purposes, creating new collective works,
  for resale or redistribution to servers or lists, or reuse of any
  copyrighted component of this work in other works.\\
  IEEE 36th Intelligent Vehicles Symposium (IV 2025) - 22-25 June, 2025.
}

\title{\LARGE \bf
Self-Supervised Pretraining for Aerial Road Extraction
}

\author{Rupert Polley$^{1}$, Sai V. A. Deenadayalan$^{1}$ and J. Marius Z\"ollner$^{1}$$^{2}$
\thanks{$^{1}$Rupert Polley, Sai V. A. Deenadayalan and J. Marius Z\"ollner are with the FZI Research Center for Information Technology, 76131 Karlsruhe, Germany
{\tt\small \{polley, deenadayalan, zoellner\}@fzi.de}}%
\thanks{$^{2}$J. Marius Z\"ollner is with the Institute of Applied Informatics and Formal Description Methods, Karlsruhe Institute of Technology (KIT), 76131 Karlsruhe, Germany
        {\tt\small marius.zoellner@kit.edu}}%
\thanks{This work is funded by the German Federal Ministry for Economic
Affairs and Climate Action within the project "nxtAIM".
}
}

\AtBeginDocument{%
  \AddToShipoutPicture*{%
    \AtPageUpperLeft{%
      \put(55, -40){
        \parbox{\dimexpr\textwidth-4pt\relax}{\raggedright \mycopyrighttext}
      }%
    }
  }
}

\begin{document}
\thispagestyle{empty}
\pagestyle{empty}

\maketitle

\begin{abstract}

Deep neural networks for aerial image segmentation require large amounts of labeled data, but high-quality aerial datasets with precise annotations are scarce and costly to produce. To address this limitation, we propose a self-supervised pretraining method that improves segmentation performance while reducing reliance on labeled data. Our approach uses inpainting-based pretraining, where the model learns to reconstruct missing regions in aerial images, capturing their inherent structure before being fine-tuned for road extraction. This method improves generalization, enhances robustness to domain shifts, and is invariant to model architecture and dataset choice. Experiments show that our pretraining significantly boosts segmentation accuracy, especially in low-data regimes, making it a scalable solution for aerial image analysis.

\end{abstract}

\section{INTRODUCTION}

An autonomous vehicle requires a robust understanding of its environment. While a human driver can navigate without prior knowledge of a location, the risk of missing critical information increases. High-Definition (HD) maps encapsulate detailed environmental information and are widely used in autonomous driving stacks to complement sensor-based perception \cite{autoware, fzistack}. These maps provide high-resolution details about lane widths, cycle paths, and sidewalks, enabling autonomous vehicles to build a more reliable model of their surroundings.
However, generating HD maps is labor-intensive and typically requires multiple passes by a vehicle equipped with an expensive sensor suite. While this approach is feasible in limited areas such as cities, scaling it globally would be prohibitively costly and time-consuming. Therefore, researchers focus on leveraging existing data sources to generate HD maps without requiring a fleet of thousands of vehicles to survey millions of kilometers. Additionally, maintaining and updating these maps requires continuous data collection, further increasing operational costs.

Aerial images captured by airplanes, satellites, or other airborne platforms serve various purposes, including agriculture, cartography, climate research, and surveillance~\cite{liu2023review}. Due to the vast number of private and state-owned enterprises producing and offering aerial imagery, extensive coverage of nearly all inhabited areas is available, making aerial images a promising candidate for HD map generation.
\newpage
However, one major drawback of aerial images is their resolution, as most providers cannot offer a ground sampling distance lower than 10~\begin{math}cm\end{math}~\cite{fischer2018towards}. Deep neural networks, commonly used for segmenting aerial images, require large datasets for effective training. The resulting segmentation masks are then processed to form HD maps. Unfortunately, labeling aerial images is time-consuming, and only a few datasets exist specifically for aerial semantic segmentation in autonomous driving. High-resolution labeled datasets are scarce and typically cover only a few square kilometers~\cite{skyscapes}, while larger datasets spanning thousands of square kilometers~\cite{DeepGlobe18, spacenet} are often limited to low-resolution imagery. Additionally, roads in these datasets are frequently labeled with centerlines rather than precise pixel-level segmentation masks.

\begin{figure*}[!ht]
\vspace*{2mm}
\centering
	\includegraphics[width=1\textwidth]{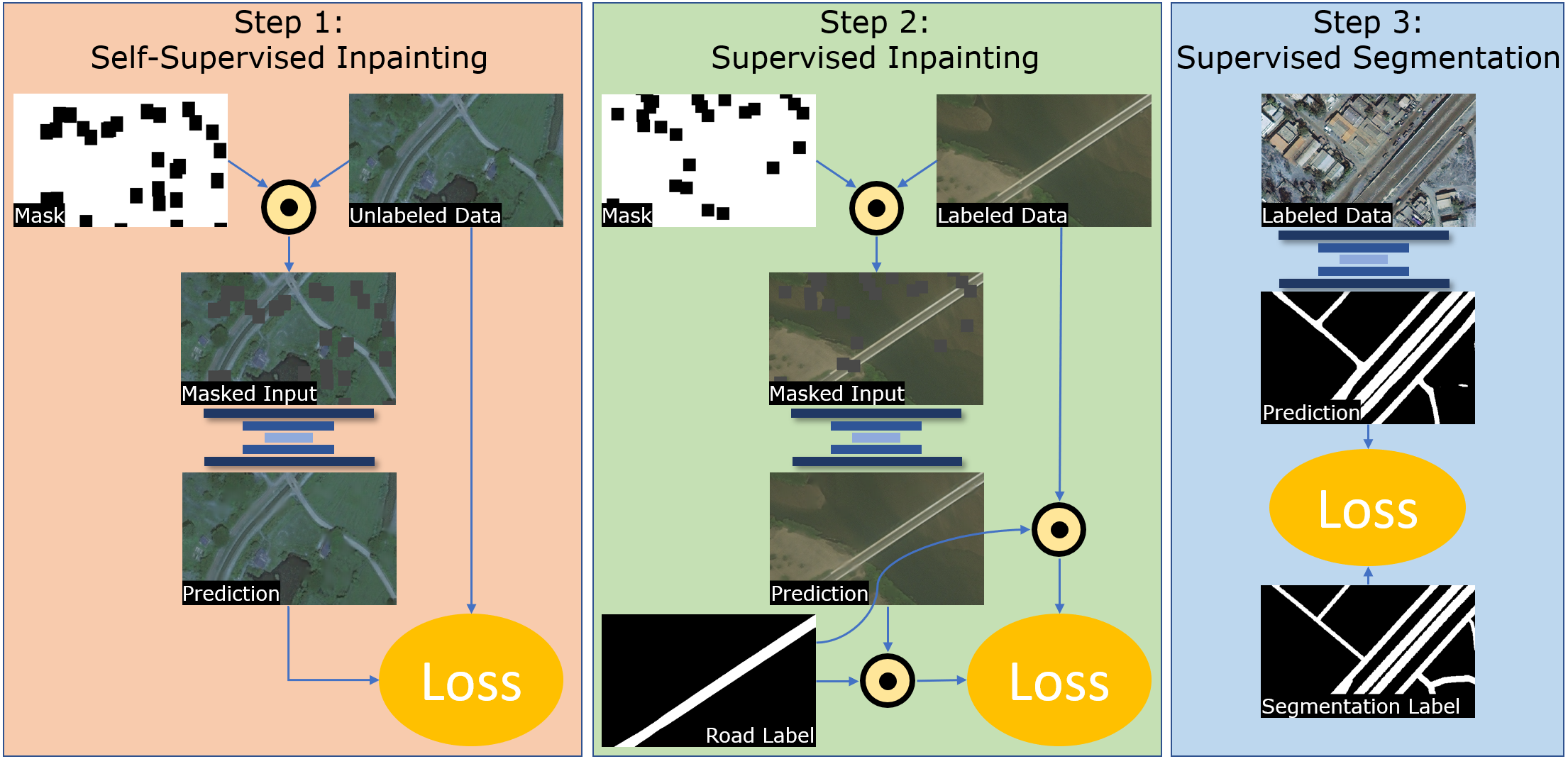}
	\caption{An overview of our proposed training method. In Step 1, a model is pretrained using inpainting on unlabeled aerial images. In Step 2, road segmentation labels are incorporated to guide the inpainting task specifically toward road structures. Finally, after pretraining on inpainting tasks, the model is fine-tuned on the road segmentation task. The symbol $\odot$ denotes the element-wise product with a binary mask.}
	\label{fig:pipeline}
\end{figure*}

To address the scarcity of labeled aerial images for semantic segmentation, we propose a self-supervised pretraining method that improves segmentation performance while reducing reliance on labeled data. Our approach introduces two key contributions:
\begin{itemize}
\item Self-supervised inpainting pretraining: A model is first trained on unlabeled aerial images using an inpainting task, enabling it to learn the underlying structures of aerial imagery without requiring labels.
\item Domain adaptation step: A second training phase bridges the domain gap between the inpainting task and the downstream road segmentation task, improving feature transferability and ensuring better alignment with the final segmentation objective.
\end{itemize}
After pretraining, the model is fine-tuned on the available labeled datasets for aerial image segmentation.
Our method reduces dependence on labeled data, improves generalization, and enhances model robustness, all while maintaining inference efficiency. Unlike conventional approaches, our method is architecture-agnostic and scalable to different datasets. We evaluate it on multiple aerial segmentation models and demonstrate consistent performance gains across different architectures, highlighting its broad applicability. An overview of our training method is shown in Figure~\ref{fig:pipeline}.

\section{RELATED WORK}

Image segmentation for aerial images is challenging due to the scarcity of pixel-level labeled data. Some studies attempt to address this issue by learning the structural and contextual information of aerial images \cite{singh2018}. Inpainting is a technique that uses contextual information to reconstruct missing regions in an image. Pretraining a network with an inpainting task could lead to better segmentation performance.

\subsection{Inpainting techniques}
Pathak et al. \cite{pathak2016context} propose a method to learn a representation of an image that captures the appearance and the semantics of visual structures by training context encoders by inpainting random missing regions of an image. Singh et al. \cite{singh2018} build upon Pathak's work by proposing a self-supervised feature learning method with an adversarial training scheme. They replace the original backbone and remove the fully connected bottleneck layer to preserve spatial context. The encoder-decoder network pretrained on inpainting is then used for semantic segmentation of aerial images. Park et al. \cite{PARK2022103111} construct an inpainting algorithm to remove unwanted vehicles in an aerial image when creating orthomosaics of aerial images. AeroRIT \cite{rangnekar2020aerorit} applies random pixel masking as a pretraining step. Large mask inpainting (LaMa) \cite{Suvorov} proposes a new architecture with fast Fourier convolutions to improve the receptive field and the inpainting performance when large areas of an image are masked out. Yu et al. \cite{Yu2020} propose an Attention GAN-based approach to reduce dependence on large labeled datasets. An attention module is added to the discriminator network to enhance its feature extraction ability. A context aggregation-based feature fusion strategy is also employed by the discriminator to aggregate contextual information. A similar pretraining objective called Masked Autoencoding is also used as a pretraining step for image classifiers \cite{he2022masked, woo2023convnext}. None of these methods address the specific task of road extraction while bridging the domain gap between a pretraining task and the downstream task.
\subsection{Aerial Image Segmentation}
Convolutional Neural Networks (CNNs) are widely used for image segmentation \cite{LongSD14}, \cite{marmanis2016semantic}, \cite{Sang2018FullyRC}. Li et al. \cite{Li2019} improve U-Net \cite{RonnebergerFB15} by integrating cascaded dilated convolutions to enhance multi-scale segmentation. Kaiser et al. \cite{KaiserWLJHS17} incorporate online non-HD maps to refine their fully convolutional network architecture. The ResUNet-a model \cite{diakogiannis2020resunet} introduces residual connections, pyramid scene parsing pooling, dilated convolutions, and a novel Dice loss to enhance performance in remote sensing applications. Benjdira et al. \cite{benjdira2019unsupervised} identify performance degradation when applying segmentation models to unseen scenes and propose a domain adaptation technique to bridge this gap. Batra et al. \cite{batra2019improved} extend road segmentation by incorporating an additional decoder branch that classifies road orientations into quantized bins. Bandara et al. \cite{bandara2022spin} introduce a Spatial and Interaction Space Graph Reasoning (SPIN) module, which captures long-range dependencies in aerial images and achieves state-of-the-art performance on the DeepGlobe \cite{DeepGlobe18} and Massachusetts Roads \cite{massa} datasets.
Commonly used datasets in the automotive and remote sensing domains include SkyScapes \cite{skyscapes}, SpaceNet \cite{spacenet}, ISPRS Potsdam and Vaihingen \cite{markus2014use}, CITY-OSM \cite{kaiser2017learning}, Air-Ground-KITTI \cite{mattyus2016hd}, and TorontoCity \cite{wang2017torontocity}. Despite advancements in segmentation models, none employ pretraining specifically optimized for aerial segmentation. Instead, they either rely on backbones pretrained on ImageNet \cite{imagenet} or initialize all weights from scratch, limiting their ability to generalize across diverse aerial imagery.

\section{METHOD}
Since most aerial road segmentation datasets contain only a limited number of labeled images, we propose an inpainting-based pretraining approach that utilizes both unlabeled and labeled data. Our architecture-agnostic method can be applied to any CNN-based model architecture and aerial segmentation dataset. The training process consists of three sequential steps, where each stage continues training the same convolutional neural network for segmentation.

\subsection{Step 1: Self-Supervised Training - Inpainting}
Since unlabeled aerial images are widely available while labeled data remains sparse, self-supervised methods are employed to enhance model performance without requiring additional annotations. The first training step consists of self-supervised inpainting, where regions of an image are removed, and a model is trained to reconstruct the missing areas. After reconstruction, the generated image is compared to the original, and a loss is computed to guide training. To remove information, a binary mask $M$ with the same resolution as the input image is created. The mask $M$ indicates which pixels are removed from the input image and which output pixels are predicted without corresponding input information. Initially, all values in $M$ are set to 1, meaning no information is removed. Then, a predefined number of random indices are selected, and the corresponding values in $M$ are set to 0. To ensure locally coherent masking, these selected locations serve as the top-left corners of square regions, where all pixels within each square are also set to 0. The size of each square is predefined, and overlapping squares can create larger masked regions.

During training, the square size gradually increases while the number of masked squares decreases, ensuring that the overall proportion of masked pixels remains constant but forms larger contiguous regions. To apply masking, $M$ is copied along the channel axis and multiplied element-wise with the input image. This element-wise multiplication removes pixel information in all locations where 
$M=0$, setting the corresponding pixel values to zero. Since the input image is normalized and augmented beforehand, a pixel with an RGB value of (0,0,0) does not necessarily represent a black pixel.

CNN-based semantic segmentation models typically follow a U-Net architecture \cite{RonnebergerFB15}, producing a segmentation mask with the same resolution as the input image, where the number of output channels corresponds to the number of segmented classes. Our method is generally agnostic to the choice of CNN architecture, requiring only minimal modifications to the segmentation head during pretraining. Specifically, in the first and second training steps, the final layer is adjusted to enable any segmentation head to predict RGB values for image reconstruction.

\textbf{Loss-Functions:} 
With the proposed inpainting technique, the model must solve two distinct tasks. The first task, referred to as \mbox{\textit{identity}}, involves learning to reproduce the input in non-masked areas. This is relatively easy for the model, as the commonly used skip connections in CNN architectures make it trivial. The second task, called \textit{fill}, requires detecting and reconstructing the masked areas. To accomplish this, the network must learn the structural patterns of aerial images. As the masked clusters grow larger during training, simple interpolation from the nearest unmasked neighboring pixels becomes insufficient. The model must instead learn to infer how roads, buildings, road markings, and other features behave in an aerial scene to accurately reconstruct the missing information. This task is significantly more challenging, as it requires a deeper understanding of aerial imagery and its long-range dependencies.
Let $L_{id}$ denote the loss function for image reproduction and $L_{fill}$ the loss function for predicting masked-out areas. Given that $M$ represents the binary mask, $X$ the unmasked input image, $O$ the model output, $\odot$ the element-wise product, and $MSE$ the mean-squared error function, the total loss in the first training step is formulated as follows:

\begin{equation}
    L_{id}^0 = MSE(O \odot M, X \odot M) \quad ,
\end{equation}
\begin{equation}
    L_{fill}^0 = MSE(O \odot (1-M), X \odot (1-M)) \quad ,
\end{equation}
\begin{equation}   
    L_{total}^0 = w_{id}^0 \cdot L_{id}^0 + w_{fill}^0  \cdot L_{fill}^0 \quad ,
    \label{Lstep1}
\end{equation}

where $w_{id}$ and $w_{fill}$ are empirically determined weights that increase the emphasis on the \textit{fill} task while reducing the influence of the \textit{identity} task. The magnitude of variation in $L_{id}$ during training is significantly larger than in $L_{fill}$, causing $L_{id}$ to dominate the optimization process in early iterations, even when weighting parameters are introduced.

Since this training step does not require labeled data, such as segmentation masks, the number of available aerial images can be significantly increased, making the approach highly scalable.

\subsection{Step 2: Supervised Training - Guided Inpainting}
\label{guidedinpainting}
Following the first inpainting step, a second training process step called guided inpainting is introduced. Due to the random placement of masking clusters and the low proportion of road pixels in aerial images, most clusters tend to obscure buildings, vegetation, and other background areas.
While buildings contain learnable structures, rural areas with fields and meadows surrounding roads provide significantly less useful structural information. Depending on the training dataset, this can cause the model to produce blurry interpolations rather than sharp reconstructions within masked regions. Since the primary goal of the initial training steps is to help the model learn the underlying structure of road networks in aerial images, we introduce a supervised guided inpainting task. This step helps bridge the domain gap between inpainting and segmentation, which is the model's final objective.
To achieve this, we use road segmentation labels to generate a binary road mask $S$ in the second training step. The mask $S$ ensures that the \textit{fill} and \textit{identity} losses are computed only for road pixels, directing the model's learning toward road structures rather than background areas.
This transforms the second training step into a supervised learning process, as it requires the binary segmentation mask $S$, which segments the aerial image into roads ($1$) and background ($0$).

\textbf{Loss-Functions:} Following the earlier notation, the new loss functions for the second training step are defined as:

\begin{equation}
    L_{id}^1 = MSE(O \odot M \odot S, X \odot M \odot S) \quad ,
\end{equation}
\begin{equation}
    L_{fill}^1 = MSE(O \odot (1-M) \odot S, X \odot (1-M) \odot S) \quad ,
\end{equation}
\begin{equation}   
    L_{total}^1 = w_{id}^1 \cdot L_{id}^1 + w_{fill}^1  \cdot L_{fill}^1 \quad .
    \label{Lstep2}
\end{equation}

By applying the binary road segmentation mask element-wise, both $L_{fill}$ and $L_{id}$ generate a loss only in regions where roads are present. Performance in all other areas is ignored, effectively redirecting the model's focus toward roads while deprioritizing buildings, vegetation, and open fields. Since roads are the most critical structures for HD map generation in later downstream tasks, this ensures that the model learns features relevant to high-precision mapping.

\subsection{Step 3: Supervised Training - Segmentation}
In the third and final training step, the model, which was previously trained with the two preceding inpainting steps, is now optimized for aerial image segmentation. This step requires restoring the segmentation head to its original architecture. All weights in the backbone, neck, and head remain unchanged. However, the final layer in the decoder head, which was modified for the inpainting tasks, is now reverted to its original configuration, enabling segmentation with a variable number of classes.


\section{IMPLEMENTATION}
Our proposed method is agnostic to both model architecture and dataset selection, making it adaptable to various segmentation frameworks. To demonstrate its versatility, we implement it in two deep learning models and apply it to different aerial datasets, ensuring its effectiveness in diverse training scenarios.
\subsection{Datasets}
Large datasets are necessary to create controlled training scenarios with varying amounts of labeled data. By systematically reducing the number of training images from a larger dataset, we simulate conditions where fewer annotations are available. This allows us to analyze the model’s performance under data-scarce conditions and observe its behavior across different dataset sizes.

\label{datasetssection}
\subsubsection{DeepGlobe} We evaluate our training method on a dataset provided by the DeepGlobe road extraction challenge \cite{DeepGlobe18}. This dataset contains pixel-wise annotated aerial images of regions in Thailand, Indonesia, and India. DeepGlobe provides 4696 training images, 1530 validation images, and 1101 test images, each with a resolution of 1024$\times$1024 and a ground sampling distance of 50~\begin{math}cm\end{math}. DeepGlobe is a single-class segmentation dataset, where roads are labeled using a binary mask, distinguishing road pixels from the background. Following \cite{batra2019improved}, the train images are cropped to 512$\times$512 with an overlap of 256 pixels, resulting in 9 cropped images for each original image. This preprocessing step creates an expanded training set of 42,264 images, which we refer to as $DG_{l}$. 

To simulate realistic scenarios where fewer labeled aerial images are available and to evaluate the effectiveness of our method, we create progressively smaller training subsets. First, we halve the number of training images to 2,348 and then apply the same cropping and overlapping strategy, resulting in a smaller training set $DG_{s}$. Halving the training set again to 1,174 images produces the smallest dataset, $DG_{n}$, which contains 10,566 cropped images. \label{DGS}
Since the DeepGlobe challenge has ended and no longer accepts submissions, we evaluate our training method using validation images, following the approach of other state-of-the-art methods  \cite{batra2019improved,bandara2022spin,zhou2018d,zhang2021stagewise,mei2021coanet}.
The first training step does not require explicit labels since it is self-supervised. This allows us to incorporate additional images where labels are unavailable. A natural choice for these additional images is the test set, as the labels for these images were never released and, therefore, cannot be used for supervised training or evaluation. In real-world applications, integrating unlabeled images with minimal domain gap is a practical approach. To simulate this, we crop and add all test images to $DG_{l}$, creating an extended self-supervised dataset, ${DG_{self-sup}}$ with 52,173 images.

\subsubsection{CITY-OSM} Additionally, we evaluate our training technique on a weakly labeled dataset named CITY-OSM \cite{kaiser2017learning}. This dataset is labeled by overlaying OpenStreetMap (OSM) data with aerial images from Google Maps, resulting in weakly annotated aerial segmentation masks. Unlike DeepGlobe, this dataset includes both roads and buildings as labeled features. The dataset covers images of Berlin, Chicago, Paris, and Zurich with a ground sampling distance of 10~\begin{math}cm\end{math} and a total area coverage of 157.7~\begin{math}km^{2}\end{math}. Following Emek \cite{github}, we split the dataset into 90\% training and 10\% validation sets. The images are cropped into patches of 256$\times$256 pixels for both training and validation using the FiveCrop augmentation method. As in the previous dataset split methodology (\ref{DGS}), we refer to the largest training set with 7,405 images as $CO_l$. We further create progressively smaller subsets by reducing the number of training samples by half ($CO_s$) and three-quarters ($CO_n$) to evaluate performance on limited labeled data.

To further assess model performance under domain shift conditions, we introduce a second CITY-OSM split, where the validation set consists exclusively of images from Berlin, while the training set includes only the other three cities. In this configuration, the model is never exposed to Berlin during training. Since 11\% of the dataset images correspond to Berlin, the total amount of training data remains comparable between Emek’s original split and our second split. To distinguish between these two configurations, we denote the training datasets without Berlin as $\bar{B}CO_l$, $\bar{B}CO_s$, and $\bar{B}CO_n$.
To enhance generalization performance, we apply image augmentation techniques, including mirroring, flipping, and 90-degree rotation, in all three training steps for both datasets.

\subsection{Models}
To demonstrate the generalizability of our training method, we apply it to multiple segmentation models. Our approach is model-agnostic, requiring only minimal modifications to adapt to different architectures.

\subsubsection{SPIN} We use SPIN RoadMapper~\cite{bandara2022spin} as a baseline for evaluating our training method on the DeepGlobe dataset, as it represents a state-of-the-art approach for road extraction and has publicly available code~\cite{githubSPIN}. For a fair comparison, we maintain the same model architecture and identical hyperparameters in the third and final training step. In the first and second training steps, an additional filter is added to the last layer of the segmentation branch to enable RGB prediction. Since the model performs binary road extraction in the third training step, this additional layer is removed.
We train the model using a batch size of 32 and optimize with SGD, using a momentum of 0.9 and a weight decay of 0.0005. \label{SPINhyper} For the first and third training steps, we apply a multistep learning rate scheduler, starting with an initial learning rate of 0.01, which is decayed by a factor of 0.1 after 50, 90, and 110 epochs. The model is trained for 120 epochs in both steps.
For the second training step, the model is trained for 40 epochs, starting with an initial learning rate of 0.001, which is decayed by a factor of 0.1 after 10, 20, and 30 epochs.
The weights \mbox{$w_{id}^0$ / $w_{id}^1$} and \mbox{$w_{fill}^0$ / $w_{fill}^1$} from Equations~\ref{Lstep1} and \ref{Lstep2} are set to 0.2 and 0.8, respectively.

\subsubsection{EmekU-Net} To demonstrate that our method generalizes beyond a specific architecture and dataset, we also conduct experiments on the $CO$ datasets using a different model architecture. Specifically, we follow Emek's U-Net implementation, which has publicly available code~\cite{github}. 
EmekU-Net is based on the model proposed in~\cite{kaiser2017learning}, which represents a state-of-the-art approach for aerial image segmentation. This architecture has been extensively used for weakly labeled datasets, making it a suitable benchmark for evaluating our pretraining method under different labeling conditions. By applying our approach to this model, we assess its effectiveness in both high-resolution, weakly labeled datasets and low-resolution, fully annotated datasets, demonstrating its robustness across different segmentation challenges. During the first and second training steps, we reuse the training hyperparameters presented in Section~\ref{SPINhyper}. In the third training step, we apply the hyperparameters suggested by Emek, optimizing with Adam and decaying the learning rate of 0.001 by a factor of 0.1 after 10, 20, and 30 epochs.
Since this model segments three classes, no architectural modifications are required between the inpainting and segmentation training steps.

\subsection{Inpainting}
\begin{figure}[!ht]
\centering
	\includegraphics[width=.48\textwidth]{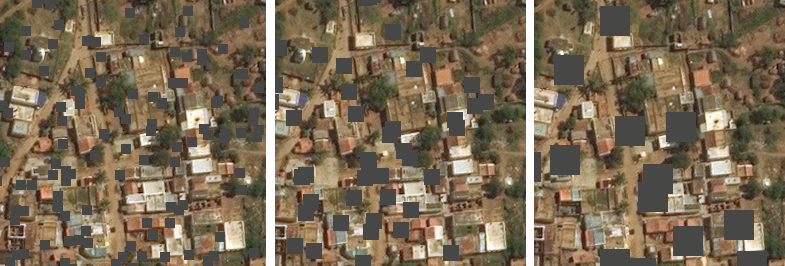}
	\caption{While the size of the clusters during training increases, the number of clusters decreases. Epochs from left to right are 10, 30, 50.}
	\label{fig:cluster}
\end{figure}

Preliminary tests indicate that inpainting with larger clusters improves performance. However, training with large clusters from the start leads to instability and failure to converge, as the task becomes significantly more challenging. To address this, we adopt a dynamic approach where cluster size is gradually increased at specific epochs.
To maintain a consistent ratio of masked pixels, we decrease the number of clusters as their size increases. This results in a high number of small clusters at the beginning of training and fewer but larger clusters toward the end. For instance, at the start of training (epoch 0), we use 100 clusters of size 10 pixels, whereas by epoch 50, the number of clusters is reduced to 11, and their size is increased to 30 pixels. The full progression of cluster sizes and counts at different epochs is shown in Table~\ref{clustertable} and illustrated in Figure~\ref{fig:cluster}.
With the selected values, the percentage of masked pixels increases only slightly over time. In the second training step, the number and size of clusters remain fixed at their final values from the first training step.

\begin{table}[ht]
\caption{Number and size of the clusters during pretraining.}
\label{clustertable}

\centering
\begin{tabular}{c|cc}
Epoch & \# Cluster & Cluster size \\ \hline
0     & 100        & 10           \\
10    & 70         & 12           \\
20    & 52         & 14           \\
30    & 50         & 15           \\
40    & 25         & 20           \\
50    & 11         & 30          
\end{tabular}
\end{table}

\section{EVALUATION}

We evaluate our training method using SPIN RoadMapper and EmekU-Net as baselines. The fundamental difference between our approach and the baselines lies in the first and second training steps, which are exclusive to our method. In contrast, the baselines are trained directly from scratch, without these additional pretraining stages. The third training step in our method is identical to the training process of the baselines, except that the baselines initialize their weights from scratch, whereas our method transfers weights from the second training step. To analyze the impact of reduced training data, we compare our method's performance against the baselines under limited data conditions. We train SPIN using the DeepGlobe datasets $DG_{l}$, $DG_{s}$, and $DG_{n}$ and EmekU-Net using both City-OSM splits $CO_l$, $CO_s$, $CO_n$ and $\bar{B}CO_l$, $\bar{B}CO_s$, $\bar{B}CO_n$. As described in Section~\ref{datasetssection}, in the $\bar{B}CO$ configuration, EmekU-Net is only trained on images from Chicago, Paris, and Z\"urich and validated on Berlin. For the self-supervised pretraining in the first step, we use $DG_{self-sup}$, $CO_l$, and $\bar{B}CO_l$. 
The resulting road and building segmentations are evaluated using the Intersection over Union (IoU) score. The performance across different dataset sizes and training strategies is summarized in Table~\ref{EmekTable1}.

\begin{table}[t]
\caption{IoU scores for SPIN and EmekU-Net on validation}
\centering
\label{EmekTable1}
\begin{tabular}{l|ccc}
 \textbf{SPIN}  & Training Images & Road IoU & Building IoU 
 \\ \hline
 $DG_{l}$ Baseline & \multirow{2}{*}{42,264} & 82.32 & ---\\
$DG_{l}$ Ours & & \textbf{82.66} & ---\\ \hline
$DG_{s}$ Baseline & \multirow{2}{*}{21,132} & 78.95 & ---\\
$DG_{s}$ Ours & & \textbf{81.80} & ---\\ \hline
$DG_{n}$ Baseline & \multirow{2}{*}{10,566} & 77.04 & ---\\
$DG_{n}$ Ours & & \textbf{80.07}  & ---\\ \hline \\
\textbf{EmekU-Net}\\ \hline
$CO_{l}$ Baseline & \multirow{2}{*}{7,405} & 42.88 & 65.02         \\
$CO_{l}$ Ours & & \textbf{52.79} & \textbf{67.33} \\ \hline
$CO_{s}$ Baseline & \multirow{2}{*}{3,700} & 25.39 & 58.89  \\
$CO_{s}$ Ours & & \textbf{52.18} & \textbf{67.03} \\ \hline
$CO_{n}$ Baseline & \multirow{2}{*}{1,850} & 19.65 & 52.90     \\
$CO_{n}$ Ours & & \textbf{51.43} & \textbf{66.41} \\ \hline 
\\ \hline
\rule{0pt}{2.5ex}$\bar{B}CO_{l}$ Baseline & \multirow{2}{*}{7,230} & 16.99 & 50.15         \\
$\bar{B}CO_{l}$ Ours & & \textbf{32.62} & \textbf{54.65} \\ \hline
\rule{0pt}{2.5ex}$\bar{B}CO_{s}$ Baseline & \multirow{2}{*}{3,615} & 16.80 & 51.65  \\
$\bar{B}CO_{s}$ Ours & & \textbf{30.82} & \textbf{54.93} \\ \hline
\rule{0pt}{2.5ex}$\bar{B}CO_{n}$ Baseline & \multirow{2}{*}{1,805} & 16.72 & 45.23     \\
$\bar{B}CO_{n}$ Ours & & \textbf{30.56} & \textbf{54.76}
\\
\end{tabular}
\end{table}

The results demonstrate that our proposed pretraining method consistently outperforms the baselines across all scenarios. Our approach is more resilient, with a smaller decline in IoU scores when training on datasets with artificially limited labeled data. This confirms that our pretraining steps are particularly effective in improving performance on datasets with few labeled samples.

Evaluating on the $\bar{B}CO$ split reveals a significant performance drop due to the domain gap when validating on imagery from an unseen city. This suggests that the baseline model struggles to generalize and fails to bridge the domain gap. In contrast, our method enhances model robustness, leading to improved domain shift performance.
However, when comparing the $\bar{B}CO$ split with $CO$, the IoU score is noticeably degraded. This indicates that further pretraining techniques are needed, as our method still underperforms compared to Emek's split.

We also assess the impact of the guided inpainting step proposed in Section~\ref{guidedinpainting}. To do so, we remove this step and instead transfer the model weights directly from the unsupervised inpainting step to train the downstream segmentation task. The resulting Road IoU scores, shown in Table~\ref{ablation}, indicate that guided inpainting enhances performance by bridging the gap between inpainting and semantic segmentation, mitigating the effects of domain shift.
The primary drawback of our training method is the increased training time. However, we argue that this is a reasonable trade-off, as the alternative would require manual annotation of additional images. Importantly, the inference time of all models remains unchanged, as our pretraining method does not modify the selected downstream model architecture or number of parameters.

\begin{table}[h]
\caption{Impact of guided inpainting on Validation Road IoU scores.}
\label{ablation}
\centering
\begin{tabular}{l|ccc}
   & Baseline & without Guided Inpainting & Proposed Method  
 \\ \hline
\rule{0pt}{2.5ex}$DG_{n} $& 77.04 & 79.18 & \textbf{80.07}   \\ \hline
\rule{0pt}{2.5ex}$CO_{n}$ & 19.65 & 48.52 & \textbf{51.43}   \\ \hline
\rule{0pt}{2.5ex}$\bar{B}CO_{n}$ &  16.72 & 26.92 & \textbf{30.56}    \\ 
\end{tabular}
\end{table}

\section{CONCLUSION}
We presented a training method to enhance the performance of automotive aerial image segmentation networks. By introducing two additional training steps, our approach improves resilience, bridges domain gaps, and enhances segmentation accuracy in state-of-the-art baselines. Experiments with artificially limited labeled data demonstrate that our method is more robust and requires fewer labeled images. Additionally, we show that it helps mitigate domain gaps, though further research is needed, as performance degradation remains when evaluating on data from an unseen city. The method proves effective for both high-resolution weakly labeled datasets and low-resolution pixel-wise annotated datasets. These qualities make it a strong candidate for training aerial segmentation networks, particularly in scenarios with limited labeled data.

\addtolength{\textheight}{-11.0cm}   





\bibliographystyle{IEEEtran}
\bibliography{root}

\begin{thebibliography}{10}
\providecommand{\url}[1]{#1}
\csname url@samestyle\endcsname
\providecommand{\newblock}{\relax}
\providecommand{\bibinfo}[2]{#2}
\providecommand{\BIBentrySTDinterwordspacing}{\spaceskip=0pt\relax}
\providecommand{\BIBentryALTinterwordstretchfactor}{4}
\providecommand{\BIBentryALTinterwordspacing}{\spaceskip=\fontdimen2\font plus
\BIBentryALTinterwordstretchfactor\fontdimen3\font minus \fontdimen4\font\relax}
\providecommand{\BIBforeignlanguage}[2]{{%
\expandafter\ifx\csname l@#1\endcsname\relax
\typeout{** WARNING: IEEEtran.bst: No hyphenation pattern has been}%
\typeout{** loaded for the language `#1'. Using the pattern for}%
\typeout{** the default language instead.}%
\else
\language=\csname l@#1\endcsname
\fi
#2}}
\providecommand{\BIBdecl}{\relax}
\BIBdecl

\bibitem{autoware}
S.~Kato, S.~Tokunaga, Y.~Maruyama, and et~al., ``Autoware on board: Enabling autonomous vehicles with embedded systems,'' in \emph{International Conference on Cyber-Physical Systems (ICCPS)}, 2018.

\bibitem{fzistack}
S.~Ochs, J.~Doll, D.~Grimm, and et~al., ``{One Stack to Rule them All: To Drive Automated Vehicles, and Reach for the 4th level},'' \emph{arXiv preprint arXiv:2404.02645}, 2024.

\bibitem{liu2023review}
X.~Liu, K.~H. Ghazali, F.~Han, and I.~I. Mohamed, ``{Review of CNN in Aerial Image Processing},'' \emph{The Imaging Science Journal}, 2023.

\bibitem{fischer2018towards}
P.~Fischer, S.~M. Azimi, R.~Roschlaub, and T.~Krau{\ss}, ``{Towards HD maps from Aerial Imagery: Robust Lane Marking Segmentation using Country-Scale Imagery},'' \emph{ISPRS International Journal of Geo-Information}, 2018.

\bibitem{skyscapes}
S.~M. Azimi, C.~Henry, L.~Sommer, A.~Schumann, and E.~Vig, ``Skyscapes fine-grained semantic understanding of aerial scenes,'' in \emph{IEEE/CVF International Conference on Computer Vision}, 2019.

\bibitem{DeepGlobe18}
I.~Demir, K.~Koperski, D.~Lindenbaum, G.~Pang, J.~Huang, S.~Basu, F.~Hughes, D.~Tuia, and R.~Raskar, ``Deepglobe 2018: A challenge to parse the earth through satellite images,'' in \emph{Conference on Computer Vision and Pattern Recognition (CVPR) Workshops}, 2018.

\bibitem{spacenet}
A.~Van~Etten, D.~Lindenbaum, and T.~M. Bacastow, ``Spacenet: A remote sensing dataset and challenge series,'' \emph{arXiv preprint arXiv:1807.01232}, 2018.

\bibitem{singh2018}
S.~Singh, A.~Batra, G.~Pang, L.~Torresani, S.~Basu, M.~Paluri, and C.~Jawahar, ``{Self-Supervised Feature Learning for Semantic Segmentation of Overhead Imagery},'' in \emph{BMVC}, 2018.

\bibitem{pathak2016context}
D.~Pathak, P.~Krahenbuhl, J.~Donahue, T.~Darrell, and A.~A. Efros, ``Context encoders: Feature learning by inpainting,'' in \emph{Conference on Computer Vision and Pattern Recognition (CVPR)}, 2016.

\bibitem{PARK2022103111}
J.~Park, Y.~K. Cho, and S.~Kim, ``{Deep learning-based UAV image segmentation and inpainting for generating vehicle-free orthomosaic},'' \emph{International Journal of Applied Earth Observation and Geoinformation}, 2022.

\bibitem{rangnekar2020aerorit}
A.~Rangnekar, N.~Mokashi, E.~J. Ientilucci, C.~Kanan, and M.~J. Hoffman, ``{AeroRIT: A New Scene for Hyperspectral Image Analysis},'' \emph{IEEE Transactions on Geoscience and Remote Sensing}, 2020.

\bibitem{Suvorov}
R.~Suvorov, E.~Logacheva, A.~Mashikhin, and et~al., ``{Resolution-Robust Large Mask Inpainting with Fourier Convolutions},'' in \emph{IEEE/CVF winter conference on applications of computer vision}, 2022.

\bibitem{Yu2020}
Y.~Yu, X.~Li, and F.~Liu, ``Attention {GANs}: Unsupervised deep feature learning for aerial scene classification,'' \emph{IEEE Transactions on Geoscience and Remote Sensing}, 2020.

\bibitem{he2022masked}
K.~He, X.~Chen, S.~Xie, Y.~Li, P.~Doll{\'a}r, and R.~Girshick, ``{Masked Autoencoders are Scalable Vision Learners},'' in \emph{Conference on Computer Vision and Pattern Recognition (CVPR)}, 2022.

\bibitem{woo2023convnext}
S.~Woo, S.~Debnath, R.~Hu, X.~Chen, Z.~Liu, I.~S. Kweon, and S.~Xie, ``{ConvNeXt V2: Co-designing and Scaling ConvNets with Masked Autoencoders},'' in \emph{Conference on Computer Vision and Pattern Recognition (CVPR)}, 2023.

\bibitem{LongSD14}
J.~Long, E.~Shelhamer, and T.~Darrell, ``{Fully convolutional networks for semantic segmentation},'' in \emph{Conference on Computer Vision and Pattern Recognition (CVPR)}, 2015.

\bibitem{marmanis2016semantic}
D.~Marmanis, J.~D. Wegner, S.~Galliani, K.~Schindler, M.~Datcu, and U.~Stilla, ``{Semantic Segmentation of Aerial Images with an Ensemble of CNNs},'' \emph{ISPRS Annals of the Photogrammetry, Remote Sensing and Spatial Information Sciences}, 2016.

\bibitem{Sang2018FullyRC}
D.~V. Sang and N.~D. Minh, ``{Fully Residual Convolutional Neural Networks for Aerial Image Segmentation},'' \emph{9th International Symposium on Information and Communication Technology}, 2018.

\bibitem{Li2019}
X.~Li, Y.~Jiang, H.~Peng, and S.~Yin, ``{An Aerial Image Segmentation Approach based on Enhanced Multi-Scale Convolutional Neural Network},'' in \emph{IEEE International Conference on Industrial Cyber Physical Systems (ICPS)}, 2019.

\bibitem{RonnebergerFB15}
O.~Ronneberger, P.~Fischer, and T.~Brox, ``{U-Net: Convolutional Networks for Biomedical Image Segmentation},'' in \emph{18th International Conference on Medical Image Computing and Computer-Assisted Intervention}.\hskip 1em plus 0.5em minus 0.4em\relax Springer, 2015.

\bibitem{KaiserWLJHS17}
P.~Kaiser, J.~D. Wegner, A.~Lucchi, M.~Jaggi, T.~Hofmann, and K.~Schindler, ``{Learning Aerial Image Segmentation from Online Maps},'' \emph{IEEE Transactions on Geoscience and Remote Sensing}, 2017.

\bibitem{diakogiannis2020resunet}
F.~I. Diakogiannis, F.~Waldner, and et~al., ``{ResUNet-a: A Deep Learning Framework for Semantic Segmentation of Remotely Sensed Data},'' \emph{ISPRS Journal of Photogrammetry and Remote Sensing}, 2020.

\bibitem{benjdira2019unsupervised}
B.~Benjdira, Y.~Bazi, A.~Koubaa, and K.~Ouni, ``{Unsupervised domain adaptation using generative adversarial networks for semantic segmentation of aerial images},'' \emph{Remote Sensing}, 2019.

\bibitem{batra2019improved}
A.~Batra, S.~Singh, G.~Pang, and et~al., ``{Improved Road Connectivity by Joint Learning of Orientation and Segmentation},'' in \emph{IEEE/CVF Conference on Computer Vision and Pattern Recognition}, 2019.

\bibitem{bandara2022spin}
W.~G.~C. Bandara, J.~M.~J. Valanarasu, and V.~M. Patel, ``{Spin Road Mapper: Extracting Roads from Aerial Images via Spatial and Interaction Space Graph Reasoning for Autonomous Driving},'' in \emph{International Conference on Robotics and Automation (ICRA)}, 2022.

\bibitem{massa}
V.~Mnih, \emph{Machine learning for aerial image labeling}.\hskip 1em plus 0.5em minus 0.4em\relax University of Toronto (Canada), 2013.

\bibitem{markus2014use}
I.~Markus~Gerke, ``{Use of the stair vision library within the ISPRS 2D semantic labeling benchmark (Vaihingen)},'' \emph{Tech. rep., ITC, University of Twente}, 2014.

\bibitem{kaiser2017learning}
P.~Kaiser, J.~D. Wegner, and et~al., ``{Learning aerial image segmentation from online maps},'' \emph{IEEE Transactions on Geoscience and Remote Sensing}, 2017.

\bibitem{mattyus2016hd}
G.~M{\'a}ttyus, S.~Wang, and et~al., ``{HD Maps: Fine-grained Road Segmentation by Parsing Ground and Aerial Images},'' in \emph{IEEE Conference on Computer Vision and Pattern Recognition}, 2016.

\bibitem{wang2017torontocity}
S.~Wang, M.~Bai, and et~al., ``Torontocity: Seeing the world with a million eyes,'' in \emph{IEEE International Conference on Computer Vision}, 2017.

\bibitem{imagenet}
J.~Deng, W.~Dong, R.~Socher, L.-J. Li, K.~Li, and L.~Fei-Fei, ``{Imagenet: A large-scale hierarchical image database},'' in \emph{Conference on Computer Vision and Pattern Recognition (CVPR)}, 2009.

\bibitem{zhou2018d}
L.~Zhou, C.~Zhang, and M.~Wu, ``{D-LinkNet: LinkNet with pretrained encoder and dilated convolution for high resolution satellite imagery road extraction},'' in \emph{IEEE Conference on Computer Vision and Pattern Recognition Workshops}, 2018.

\bibitem{zhang2021stagewise}
L.~Zhang, M.~Lan, J.~Zhang, and D.~Tao, ``{Stagewise Unsupervised Domain Adaptation with Adversarial Self-Training for Road Segmentation of Remote-Sensing Images},'' \emph{IEEE Transactions on Geoscience and Remote Sensing}, 2021.

\bibitem{mei2021coanet}
J.~Mei, R.-J. Li, W.~Gao, and M.-M. Cheng, ``Coanet: Connectivity attention network for road extraction from satellite imagery,'' \emph{IEEE Transactions on Image Processing}, 2021.

\bibitem{github}
A.~Emek, ``{Aerial-Segmentation},'' \url{https://github.com/alpemek/aerial-segmentation}, 2020.

\bibitem{githubSPIN}
C.~Bandara, ``{SPIN Road Mapper: Extracting Roads from Aerial Images via Spatial and Interaction Space Graph Reasoning for Autonomous Driving},'' \url{https://github.com/wgcban/SPIN\_RoadMapper}, 2022.

\end{thebibliography}
\end{document}